\newcommand{\datewritten}{December 8\textsuperscript{th}, 2019}

\documentclass [10pt,twoside,english]{article}
\usepackage[utf8]{inputenc} %
\usepackage[T1]{fontenc}    %
\usepackage{booktabs}       %
\usepackage{amsfonts}       %
\usepackage{nicefrac}       %
\usepackage{microtype}      %
\usepackage[letterpaper]{geometry}
\usepackage{fancyhdr}
\usepackage{graphicx}

\usepackage[hyphens]{url}
\usepackage[hidelinks]{hyperref}
\hypersetup{breaklinks=true}

\usepackage{natbib}
\def\bibsep{\smallskipamount}%
\def\newblock{\ }%
\bibpunct[, ]{[}{]}{,}{n}{}{,}%

\usepackage{amsthm}
\usepackage{amsmath}
\usepackage{amssymb}
\usepackage{epstopdf}
\usepackage{setspace}
\usepackage{wrapfig}
\usepackage{babel}
\usepackage{url}

\usepackage{color}
\usepackage{subfigure}
\usepackage{manyfoot}
\usepackage{caption}
\usepackage{graphicx}
\usepackage{tabularx}

\usepackage{enumerate}
\usepackage{amssymb}
\usepackage{amsbsy}
\usepackage{amsmath}
\usepackage{amsthm}
\usepackage{amsfonts}
\usepackage{latexsym}
\usepackage{graphicx}
\usepackage{xifthen}
\usepackage{xspace}
\usepackage{bbm}

\newcommand{\eg}{{\it e.g.}}

\newcommand{\mc}[1]{\mathcal{#1}}
\newcommand{\mbf}[1]{\mathbf{#1}}

\newcommand{\defeq}{:=}

\newcommand{\what}[1]{\widehat{#1}} %

\newcommand{\indic}[1]{\mbf{1}\left\{#1\right\}} %

\newcommand{\R}{\mathbb{R}}
\makeatletter
\long\def\@makecaption#1#2{
  \vskip 0.8ex
  \setbox\@tempboxa\hbox{\small {\bf #1:} #2}
  \parindent 1.5em  %
  \dimen0=\hsize
  \advance\dimen0 by -3em
  \ifdim \wd\@tempboxa >\dimen0
  \hbox to \hsize{
    \parindent 0em
    \hfil
    \parbox{\dimen0}{\def\baselinestretch{0.96}\small
      {\bf #1.} #2
    }
    \hfil}
  \else \hbox to \hsize{\hfil \box\@tempboxa \hfil}
  \fi
}
\makeatother

\renewcommand{\P}{\mathbb{P}} %
\newcommand{\simiid}{\stackrel{\rm i.i.d.}{\sim}}

\newtheorem*{claim*}{Claim}

\newenvironment{proof-sketch}{\noindent{\bf Sketch of Proof}
  \hspace*{1em}}{\qed\bigskip\\}
\newenvironment{proof-idea}{\noindent{\bf Proof Idea}
  \hspace*{1em}}{\qed\bigskip\\}

\newenvironment{proof-of-claim}{\noindent{\bf Proof of Claim}
  \hspace*{1em}}{\qed\bigskip\\}

\newenvironment{proof-of-lemma}[1][{}]{\noindent{\bf Proof of Lemma {#1}}
  \hspace*{1em}}{\qed\bigskip\\}
\newenvironment{proof-of-proposition}[1][{}]{\noindent{\bf
    Proof of Proposition {#1}}
  \hspace*{1em}}{\qed\bigskip\\}
\newenvironment{proof-of-theorem}[1][{}]{\noindent{\bf Proof of Theorem {#1}}
  \hspace*{1em}}{\qed\bigskip\\}
\newenvironment{inner-proof}{\noindent{\bf Proof}\hspace{1em}}{
  $\bigtriangledown$\medskip\\}
\newenvironment{proof-attempt}{\noindent{\bf Proof Attempt}
  \hspace*{1em}}{\qed\bigskip\\}

\newcounter{example}

\newenvironment{example*}[1][]{
  \ifthenelse{\isempty{#1}}{%
    \noindent \textbf{Example:}\hspace*{.05em}
  }{%
    \noindent \textbf{Example} ({#1})\textbf{:}\hspace*{.05em}
  }
}{%
  $\clubsuit$ \bigskip
}

\newcounter{remark}

\newenvironment{remark*}[1][]{
  \ifthenelse{\isempty{#1}}{%
    \noindent \textbf{Remark:}\hspace*{.05em}
  }{%
    \noindent \textbf{Remark} ({#1})\textbf{:}\hspace*{.05em}
  }
}{%
  $\diamondsuit$ \bigskip
}

\newcommand{\obj}{f}  %
\newcommand{\thresh}{\gamma}

\newcommand{\normal}{\mathsf{N}}  %

\usepackage{algorithm}
\usepackage{algpseudocode}
\usepackage{algcompatible}
\usepackage{placeins}

\definecolor{matlab_blue}{rgb}{0,    0.4470,   0.7410}
\definecolor{matlab_red}{rgb}{0.8500,    0.3250,    0.0980}
\definecolor{matlab_yellow}{rgb}{0.9290,    0.6940,    0.1250}
\definecolor{matlab_purple}{rgb}{0.4940,    0.1840,    0.5560}
\definecolor{matlab_green}{rgb}{0.4660,    0.6740,    0.1880}

\newcommand\tp{\textsc{TestPilot}}
\newcommand\op{\textsc{OpenPilot}}

\DeclareNewFootnote{A}
\DeclareNewFootnote{B}

\let\footnoteR\footnoteB
\let\footnote\footnoteA

\makeatletter
\newcommand\footnoteref[1]{\protected@xdef\@thefnmark{\ref{#1}}\@footnotemark}

\numberwithin{equation}{section}
\numberwithin{figure}{section}
\numberwithin{table}{section}

\date{}
\makeatother

\title{Efficient Black-box Assessment\\of Autonomous Vehicle Safety}

\begin{document}
\abovedisplayskip 5 pt
\belowdisplayskip 5 pt

\begin{center}
	$\;$\\$\;$\\$\;$\\$\;$\\$\;$\\$\;$\\
  {\LARGE Efficient Black-box Assessment} \\
  \vspace{0.4cm}
  {\LARGE of Autonomous Vehicle Safety} \\
  \vspace{1cm} {\Large Justin Norden\footnoteR{\label{foot:equal}Equal contribution}~~~Matthew O'Kelly\footnoteref{foot:equal}~~~Aman Sinha\footnoteref{foot:equal}}\\
  \vspace{0.5cm}
{\large \texttt{\{justin, matt, aman\}@trustworthy.ai}\\
        \vspace{0.5cm}
        Trustworthy AI\\
        Palo Alto, CA, USA\\
        \vspace{0.5cm}
        \datewritten\\
        \vspace{1cm}
}
\end{center}
\thispagestyle{empty}

\begin{abstract}

While autonomous vehicle (AV) technology has shown substantial progress, we still  lack tools for rigorous and scalable testing. Real-world testing, the \textit{de-facto} evaluation method, is dangerous to the public. Moreover, due to the rare nature of failures, billions of miles of driving are needed to statistically validate performance claims. Thus, the industry has largely turned to simulation to evaluate AV systems. However, having a simulation stack alone is not a solution. A simulation testing framework needs to prioritize which scenarios to run, learn how the chosen scenarios provide coverage of failure modes, and rank failure scenarios in order of importance. We implement a simulation testing framework that evaluates an entire modern AV system as a black box. This framework estimates the probability of accidents under a base distribution governing standard traffic behavior. In order to accelerate rare-event probability evaluation, we efficiently learn to identify and rank failure scenarios via adaptive importance-sampling methods. Using this framework, we conduct the first independent evaluation of a full-stack commercial AV system, Comma AI's {\op}.

\end{abstract} \clearpage

\section{Introduction}
The risks of autonomous vehicle (AV) technologies remain largely unknown. A handful of fatal accidents involving AVs highlight the importance of testing whether the system, as a whole, can safely interact with humans. Unfortunately, the most straightforward way to test AVs---in a real-world environment---requires prohibitive amounts of time due to the rare nature of serious accidents. Specifically, a recent study demonstrates that AVs need to drive ``hundreds of millions of miles and, under some scenarios, hundreds of billions of miles to create enough data to clearly demonstrate their safety''~\citep{KalraPa16}. To reduce the time, cost, and dangers of testing in the real world, the field has largely turned to simulation-based testing methods.%

Most popular strategies for simulation-based testing are built upon decades of techniques developed for validating traditional software systems. Nevertheless, the complexities of modern AV pipelines push the boundaries of these approaches and draw out significant limitations (see Section \ref{sec:related} for specific shortcomings of each technique).
We broadly illustrate these limitations by outlining desiderata for an AV-testing framework to enable continuous integration and improvement.
The term \emph{policy} is used in the rest of this paper to describe an AV's behavior, which is defined as a mapping from observations to actions. 
In order to tractably measure and improve performance of an AV policy, a testing framework should have the following features:
\parskip 0 pt
\begin{itemize}
\itemsep=0pt
\item \textbf{Safety}: Real-world public road testing puts lives in danger. Testing must begin in a controlled setting (\eg~simulation or closed courses).
\item \textbf{Efficiency}: Industrial AV systems have regular updates. Safety testing must efficiently find failures and compare to previous performance at the pace of development (\eg~daily).
\item \textbf{Coverage}:  Understanding coverage of test scenarios is essential to evaluating the true performance of an AV policy. Finding a one-off adversarial situation amounts to finding a minimum in the landscape of AV performance. In contrast, coverage means finding all high-likelihood failure scenarios, or the area of the performance landscape which is below an acceptable level. This latter quantity is required for accurate estimates of risk.

\item \textbf{Black-box interaction}: Testing a whole system as a black box maintains its integrity and confidentiality. Integrity ensures we capture as accurately as possible the true risk of an AV policy. Combining tests of individual components does not yield this result, and inserting a scaffolding for testing interfaces between components can fundamentally change the nature of failures. Confidentiality of internal system behavior ensures there is no leakage of proprietary knowledge about the AV policy's design; this is especially important to establish cooperation of AV manufacturers with regulators or third-party testers.
\item \textbf{Adaptivity}: Since AV policies have regular updates, testing on a static set of previously-defined scenarios or failure modes results in overfitting and false confidence in safety. Adaptive testing automatically modifies test scenarios as AV policies (and their failure modes) change.
\item \textbf{Unbiasedness}: Unbiasedness implies that estimates of risk are not systemically warped by priors, previous performance, or artifacts of the testing environment that deviate from true operating conditions.
\item \textbf{Prioritized results}: For actionable insights, safety testing must rank and prioritize failures. In particular, the relative importance of modes within the long tail of failures should be quantified.
\end{itemize}
\parskip 5pt

Now we outline our approach to satisfying these requirements. We utilize a probabilistic paradigm---which we describe as a \emph{risk-based framework}~\citep{okelly2018scalable}---where the goal is to evaluate the
\emph{probability of an accident} under a base distribution representing
standard traffic behavior. By assigning learned probabilities to
environmental states and agent behaviors, the risk-based framework considers
performance of the AV policy under a data-driven model of the world. Accompanying this framework, we develop an efficient method to adaptively find failures, return unbiased estimates of risk, and rank failures according to the learned probabilistic model. We implement this framework in a black-box simulation environment that requires minimal interaction with the policy under test.

With these three components---the risk-based framework, an efficient failure-search algorithm, and a black-box simulation environment---we perform the first independent evaluation of a commercially-available AV policy, Comma AI's {\op}. {\op} is an open-source highway driver-assistance system that includes industry-standard components such as deep-learning based perception and model-predictive control. It has driven more than 10 million miles on public highways~\citep{comma19twitter}. %
By evaluating our approach on this policy, we demonstrate the viability of our framework as a tool for rigorous risk quantification and scalable continuous integration (CI) of modern AV software.

In Section~\ref{sec:approach}, we formally define our risk-based framework. Since we aim to test policies with reasonable performance, our search for failures is a \emph{rare-event simulation} problem~\citep{AsmussenGl07}; we illustrate an adaptive nonparametric importance-sampling approach to search for rare events efficiently and in an unbiased manner. In Section \ref{sec:systems}, we describe a scalable, distributed simulation system. Notably, {\op}'s codebase operates on a bespoke distributed architecture in Python and C++, uses hardware acceleration, and has never been ported to a simulation environment. We describe how this integration was performed allowing deterministic, synchronized execution of the otherwise real-time, asynchronous software. Finally, in Section~\ref{sec:experiments} we perform a case study analyzing failures of {\op} over highway scenarios. Our approach estimates \op's failure rate of 1 in 1250 miles by using two orders of magnitude fewer simulations than standard methods.

\subsection{Related work}\label{sec:related}

A simplistic approach to using simulation for AV policy testing is through a static set of predetermined challenges (\eg~\citet{schram2013implementation}). %
However, the repeated use of a small number (relative to the variations present in reality) of test-cases often leads to overfitting; passing these tests gives no guarantees of generalization to real-world performance. 
``Fuzzing''~\citep{hutchison2018robustness,dreossi2019verifai} attempts to improve generalization performance by adding slight variations to predetermined scenarios. Fuzzing considers coverage of code execution paths as opposed to realism or likelihood of failure modes~\citep{koopman1999comparing, wagner2015philosophy, drozd2018fuzzergym}. As such, this approach is best applied when the space of allowable perturbations is bounded and tractable to cover exhaustively, such as for a subcomponent of a system. Fuzzing to test AV performance as a whole system is challenging for two reasons. First, exhaustively covering a high-dimensional parameter space requires assumptions that can bias results (see \eg~\citet{wagner2015philosophy}). Second, the space of allowable perturbations for the whole system is usually unbounded or only loosely bounded by physical/biological constraints; these spaces are best described by a probabilistic distribution that weights the importance of perturbations by their likelihoods. For example, the space of \emph{all physically attainable} human-driving behaviors contains many failure scenarios that are extremely improbable in the real world. Regardless of the code executions they spawn, these improbable failures are much less relevant to understanding safety than higher-likelihood failures. Finally, it is not obvious how to translate performance on a fuzzed testing set to a statistical claim on generalization performance.

Formal verification of an AV policy's ``correctness''~\citep{KwiatkowskaNoPa11,Althoff2014, SeshiaSaSa15, apex_SAE16} is a potential candidate to reduce the intractability of empirical validation. However, formal verification is exceedingly difficult because driving policies are subject to crashes caused by other stochastic agents, and thus policies can never be completely safe under all scenarios~\citep{shalev2017formal}. Furthermore, verification of systems with deep neural networks is an active research question (see \eg~\citet{AroraBhGeMa14, CohenShSh16, BartlettFoTe17}, and \citet{SinhaNaDu17}) and the problem is NP-hard for most modern deep-learning architectures with non-smooth activations (\eg~ReLUs)~\citep{KatzBaDiJuKo17, TjengTe17}. Sometimes, verification of individual subcomponents is tractable, but there is no guarantee that this translates to overall safety when all subsystems operate together. One avenue to reduce intractability is to systematically rule out scenarios where the AV is not at fault \citep{shalev2017formal}. Unfortunately, this is a task subject to logical inconsistency, combinatorial growth in specification
complexity, and subjective assignment of fault~\citep{okelly2018scalable}.

Falsification~\citep{tuncali2016utilizing}---finding \emph{any} failure scenario in the parameter space---is an obvious relaxation of formal verification. Similar to fuzzing, this approach does not account for failure likelihoods. As such, falsification algorithms often return collections of exotic adversarial scenarios, which do not yield actionable insights into prioritizing system updates. Probabilistic variants of falsification~\citep{lee2018adaptive, koren2018adaptive} attempt to find the highest-likelihood failures, but they require a white-box Markov decision process (MDP) model of the system; defining this model can be challenging for complex systems, and its behavior may deviate from that of the real system.
\section{Proposed approach}\label{sec:approach}
Below, we formally define our risk-based framework. This framework prioritizes failures based on likelihood and requires minimal, black-box access to the AV policy under test. Additionally, we illustrate the computational challenges of finding failures and then describe our approach which is efficient, adaptive, and unbiased. Finally, we outline various approaches to building a generative model over behaviors for environmental agents that interact with the AV policy under test.

\subsection{Risk-based framework}\label{sec:framework}

We let $P_0$ denote the base distribution that models standard
traffic behavior and $X \sim P_0$ be a realization of the simulation
(\eg~weather conditions and driving policies of other agents). For a continuous 
objective function $f: \mc{X} \to \R$ that measures ``safety''---so that low
values of $f(x)$ correspond to dangerous scenarios---our goal is to evaluate
the probability of a dangerous event
\begin{equation}
p_{\gamma} \defeq \P_0(\obj(X) < \gamma),
\end{equation}
for some threshold $\gamma$. Examples of this objective function $f$ include simple performance measures such as the (negative) maximum magnitude acceleration or jerk during a simulation. More sophisticated measures include the minimum distance to any other object during a simulation or the minimum time-to-collision (TTC) during a simulation. The objective can even target specific aspects of system performance. For example, to stress-test a perception module, the objective can include the (negative) maximum norm of off-diagonal elements of the perception system's confusion matrix.

Importantly, this approach is agnostic to the complexity of the AV policy under test, as we view it as a black box: given simulation parameters $X$, the black box evaluates $f(X)$ by running a simulation. In particular, analysis of systems with deep-learning based perception systems is straightforward in this paradigm, whereas such systems render formal verification methods intractable. Moreover, this framework requires the output of only a single floating point number $f(X)$ from a simulation. This means that neither the simulation nor ego-policy codebases needs to be modified for testing apart from changes already present to run the policy in a simulated environment rather than on a real vehicle. Furthermore, potentially sensitive information about the precise meaning of $X$ within the simulation environment and the AV's decision-making behavior remains confidential from the testing algorithm. One limitation of this approach is the inherent bias introduced by the single objective $f$. To alleviate this shortcoming, one can also consider extensions that evaluate multiple risk metrics or a stochastic framework where $f$ itself is sampled from a probabilistic distribution. The latter is particularly useful for nondeterministic AV policies. Of course, these extensions also leak more information about the ego policy.

\paragraph*{Inefficiency of naive Monte Carlo in rare-event simulation}\label{sec:risk}
For AV policies which approach or exceed human-level performance, adverse events are rare and the
probability $p_{\gamma}$ is close to $0$. Thus, estimating $p_{\gamma}$ is a rare-event
simulation problem (see~\citet{AsmussenGl07} for an overview of this topic).
For rare probabilities $p_{\gamma}$,
the naive Monte Carlo method is prohibitively inefficient. Specifically, for a sample
$X_i \simiid P_0$, naive Monte Carlo uses
$\what{p}_{N, \thresh} \defeq \frac{1}{N} \sum_{i=1}^N \indic{\obj(X_i) <
	\thresh}$ as an estimate of $p_{\thresh}$. When $p_{\thresh}$ is small,
relative accuracy is the appropriate measure of performance. From the central
limit theorem, relative accuracy follows
\begin{equation*}
\left| \frac{\what{p}_{N, \thresh}}{p_{\thresh}} - 1\right|
\stackrel{\rm dist}{=}
\sqrt{\frac{1-p_{\thresh}}{N p_{\thresh}}} \left|Z\right|
+ o(1/\sqrt{N})
~~ \mbox{for}~ Z \sim \normal(0, 1).
\end{equation*}
In order to achieve $\epsilon$-relative accuracy, we need
$N \gtrsim \frac{1 - p_{\thresh}}{p_\thresh \epsilon^2}$ simulations. For light-tailed $\obj(X)$, then
$p_{\thresh} \propto \exp(-|\thresh|)$ so that the required sample size for 
naive Monte Carlo grows exponentially in $|\thresh|$.  In the next
section, we use adaptive sampling techniques that sample unsafe events more
frequently to make the evaluation of $p_{\gamma}$ tractable.

\subsection{Nonparametric importance-sampling approach}\label{sec:AIS}
To address the shortcomings of naive Monte Carlo in estimating
rare event probabilities $p_{\gamma}$, we use a multilevel splitting method~\citep{kahn51, botev2008efficient, guyader2011simulation, webb2018statistical} combined with adaptive importance sampling (AIS)~\citep{AsmussenGl07}. Multilevel splitting decomposes the rare-event probability $p_{\gamma}$
into conditional probabilities with interim threshold levels
$\infty=:L_0 > L_{1}\ldots >L_K:=\thresh$, 
\begin{align}
\label{eqn:splitting}
\P_0(\obj(X) < \thresh) &= \prod_{k=1}^K \P(\obj(X) < L_k | \obj(X) < L_{k-1}).
\end{align}
This decomposition introduces a product of non-rare probabilities
that are easy to estimate individually. Markov chain Monte Carlo (MCMC) is used to estimate each
term; obtaining an accurate estimate is tractable as long as consecutive
levels are close and the conditional probability is
therefore large. Intuitively, the splitting method iteratively steers samples $X_i$ to
the rare set $\{X | \obj(X) < \thresh\}$ through a series of supersets
$\{X | \obj(X) < L_k\}$ with decreasing objective values
$L_k> L_{k+1} \ge \gamma$.
{
\begin{algorithm}[!t]
	\caption{Basic adaptive multilevel splitting (AMS)}
	\label{alg:AMS}
	\begin{algorithmic}[1]
		\STATE {\bfseries Input:} Discarding factor $\delta$, \# of MCMC steps $T$, termination threshold $\thresh$, $N$ initial samples $X_i \simiid P_0$
		\STATE Initialize $L_0\gets\infty,\;\log(\hat p_{\thresh})\gets0,\;k\gets0$
		\WHILE{$L_k> \thresh$}
		\STATE $k\gets k+1$
		\STATE Evaluate and sort $f(X_i)$ in decreasing order
		\STATE $L_{k}\gets\max\{\gamma, f(X_{(\delta N)})\}$
		\STATE $\hat P_k \gets \frac{1}{N} \sum_{i=1}^N \indic{\obj(X_i) < L_k}.$
		\STATE $\log(\hat p_{\thresh}) \gets \log(\hat p_{\thresh}) + \log(\hat{P}_{k})$
		\STATE Discard $X_{(1)},\ldots, X_{(\delta N)}$ and resample with replacement from $X_{(\delta N +1)},\ldots, X_{(N)}$
		\STATE Apply $T$ MCMC transitions to each of $X_{(1)},\ldots, X_{(\delta N)}$
		\ENDWHILE
	\end{algorithmic}
\end{algorithm}
}

Since it is a priori unclear how to choose the levels $\infty=:L_0 > L_{1}\ldots >L_K:=\thresh$, adaptive variants of multilevel splitting choose these levels online. An elementary instantiation of adaptive multilevel splitting (AMS) chooses each intermediate level $L_k,\;k\le K-1$ by fixing $\delta \in (0,1)$, the fraction of samples that is discarded at each intermediate step (Algorithm \ref{alg:AMS}). The discarding fraction $\delta$ trades off two dueling objectives; for small values, each term in the product~\eqref{eqn:splitting} is large and hence easy to estimate by MCMC; for large values, the number of total iterations $K$ until convergence is reduced and more samples ($\delta N$) at each iteration can be simulated in parallel.

The nonparametric AMS approach complements parametric AIS methods such as the cross-entropy method~\citep{RubinsteinKr04}. Parametric AIS methods postulate models for the form of the optimal importance-sampling distribution $P_0(\cdot | f(\cdot) < \gamma)$, and they require iterative optimization of objectives with likelihood ratios. Both of these issues can be challenging in high dimensions, wherein failure regions are often discontinuous and likelihood ratios are numerically unstable. AMS does not require optimization over likelihood ratios nor does it need to postulate models for the failure region(s). On the other hand, the ``modes'' of failure AMS discovers are limited by the number of samples and the mixing properties of the MCMC sampler employed. Furthermore, contrary to parametric AIS methods, AMS has several convergence guarantees including those for bias, variance, and runtime (see~\citet{brehier2015analysis} for details). Notably, AMS is unbiased and has relative variance which scales as $\log(1/{p_{\gamma}})$ as opposed to $1/p_{\gamma}$ for naive Monte Carlo (cf. Section \ref{sec:risk}). Intuitively, AMS computes $O(\log(1/{p_{\gamma}}))$ independent probabilities, each with variance independent of $p_{\gamma}$.

The nonparametric nature of AMS allows efficient discovery of disparate failure modes. However, AMS does not generate an efficient sampler over these modes for further investigation. Instead, we learn a generative model of failures from the empirical distribution of failure modes discovered by AMS. Specifically, we use normalizing flows as our class of generative models.\footnote{As with any importance sampling technique, estimates of risk from our sampler are unbiased.} 
Normalizing flows transform a base distribution (usually a standard normal distribution) through a series of deterministic, invertible functions. Using the standard change-of-basis transform, the density $\rho(g(x))$ for some invertible map $g: \mathcal{X} \to \mathcal{Y}$ is given by:
\begin{align*}
	y = g(x) \implies \rho(y) = \rho(g^{-1}(y)) \cdot |\text{det} J(g^{-1}(y))|,
\end{align*}
where $J(\cdot)$ denotes the Jacobian. Composing a sequence of such transforms $g_1(x), g_2(g_1(x)) \dots$ gives expressive transformations of the base density.~\citet{rezende2015variational} describe a modern form of the approach which ensures the Jacobian is triangular, rendering the determinant computation efficient. Since each transform is parameterized by trainable weights, one can learn a distribution with this architecture of composed transformations by maximizing the log-probability of the observed data in the transformed distribution. Further enhancements to this approach can be found in \citet{kingma2016improved} and \citet{papamakarios2017masked}.

\subsection{Data-driven generative modeling of agent behaviors for $P_0$}
To implement our risk-based framework, we first learn a base distribution
$P_0$ of nominal traffic behavior. Although this is a challenging open problem, there are a variety of approaches in the literature to do so via imitation learning~\citep{Russell98,RossBa10,RossGoBa11,HoEr16} on data of real human drivers. Behavior cloning~\citep{bansal2018chauffeurnet} simply treats imitation learning as a standard supervised learning problem, whereas reinforcement-learning techniques such as generative adversarial imitation learning~\citep{HoEr16} improve generalization performance~\citep{RossBa10,RossGoBa11}. Recent advances in imitation-learning techniques specifically for driving vehicles can be found in \citet{kuefler2018burn, codevilla2018end, rhinehart2018deep, codevilla2019exploring}, and \citet{rhinehart2019precog}. In this paper, we use an agent modeling framework previously validated on highway scenarios in the NGSim dataset~\citep{Ngsim08} by \citet{KueflerMoWhKo17} and~\citet{okelly2018scalable}.

\section{Systems}\label{sec:systems}
In this section we describe the integration of a widely-deployed autonomous vehicle system, {\op}~\citep{openpilot}, and a popular simulation environment, CARLA~\citep{dosovitskiy2017carla}, with the theoretical framework of the previous section.\footnote{The authors of this paper are not affiliated with Comma AI.}

\paragraph*{Creating a test-harness for \op}
{\op} is a SAE Level 2~\citep{sae3016} driver-assistance system designed to be integrated with Comma AI's proprietary camera hardware %
and a customized software platform based on the Android operating system. %
At the time of writing, {\op} has been installed on more than 4500 vehicles and driven more than 10 million miles~\citep{comma19twitter}. %
A lack of publicly-discussed incidents suggests that it is a capable system, but no rigorous statistics have been released regarding its safety or performance.

\begin{figure}[!!t]
	\centering
	\includegraphics[width=0.85\columnwidth]{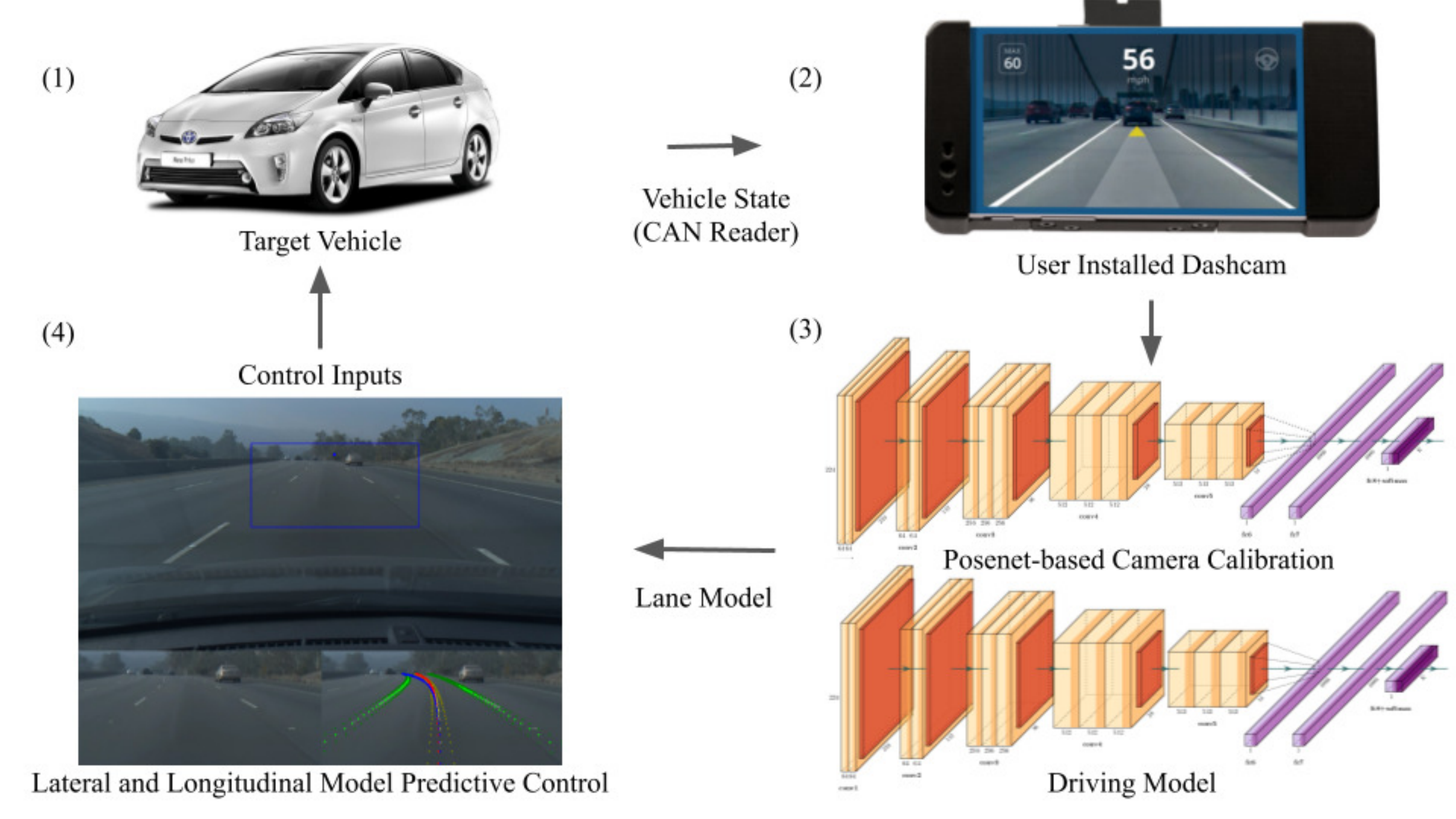}%
	\caption[]{\label{fig:op} Components of {\op}.}%
\end{figure}

Unlike end-to-end systems, {\op} contains a relatively standard set of modules for perception, planning, and control. Figure~\ref{fig:op} gives a high-level overview of the four primary components of {\op}: (1) State information is collected from the target vehicle's steering, throttle, and brake modules as well as processed-radar outputs. (2) The EON dash camera contains a camera, processor, and an interface to the vehicle. (3) The collected sensor observations, including images from the camera, are sent to the perception system where camera calibration is performed and a model of the driving environment is created. (4) Finally, the driving environment data is fed to a model-predictive control (MPC) system which outputs a trajectory to be tracked by a low-level PID controller.

Despite public availability of the source code, the majority of the system has not previously been compiled for x86 hardware, even, to the best of our knowledge, by Comma AI. we now describe {\tp}, a minimally invasive fork of {\op} which enables deterministic execution of all components on x86 hardware with support for simulator-time synchronization and optional GPU acceleration within a distributed, containerized environment.

A primary challenge associated with testing {\op} in a simulated environment is the tight integration between the neural networks in the perception system and the EON hardware. At the time of writing, the networks are distributed in binary form and use the Snapdragon Neural Processing Engine (SNPE) accelerator~\citep{ignatov2018ai}. The first network, ``posenet,'' infers the motion of the camera sensor (rather than to localize the vehicle in a map as in~\citet{kendall2015posenet}). The predicted motion allows {\op} to estimate the three-dimensional locations of objects detected by the ``driving\_model'' network. 
{\tp} utilizes an SNPE SDK feature which enables the execution of network binaries using OpenCL. Additionally, we create a ``fake'' camera driver which inputs simulated, recorded, or real images from any source %
into an OpenCL memory buffer. %

Deploying the planning and control modules together with the perception pipeline in a simulation environment requires further refinements. The perception and planning elements use a real-time clock to stamp vehicle-state messages. However, the simulation clock does not necessarily match wall-clock time. To address this problem, {\tp} receives a clock signal from the simulator. Furthermore, {\op} heavily utilizes asynchronous message passing; this induces non-determinism in the execution order of the control stack. In order to improve the repeatability of the experiments in Section \ref{sec:experiments}, {\tp} forces sequential execution of the control components. Lastly, {\op} utilizes a PID controller operating at 100Hz to actuate steering and throttle mechanisms and track the desired state (updated by the planning module at 20Hz). To avoid significant time costs and execute the simulator at 20Hz, {\tp} sends the planning module's desired state to the simulator, and we perform tracking on the simulator side.

There are several differences between the experiments conducted using {\tp} below and what a user might experience in an {\op}-equipped vehicle. First, {\op} contains a driver-monitoring system that disengages when a user diverts attention from the road. We do not model these disengagement events in our experiments. Second, most {\op}-equipped vehicles utilize a radar system to aid vision-based vehicle detection. At the time of writing, open-source vehicle simulators such as CARLA do not contain a radar sensor model. Thus, it is possible that a radar-equipped {\op} vehicle may be able to avoid some of the crashes reported in Section~\ref{sec:experiments}.

\paragraph*{Running simulations at scale}\label{sec:simulator}
To support running simulations at scale, we package {\tp} modules and CARLA simulation servers as containerized applications within a standard container orchestration scheme on commodity cloud servers. Our adaptive sampling algorithm coordinates jobs via the open-source remote-procedure-call framework gRPC~\citep{grpc}. For some context, we simulated roughly 40,000 miles to generate the results in Section~\ref{sec:experiments} comparing naive Monte Carlo with our proposed approach. As we show below, our approach requires orders of magnitude fewer simulations to estimate risk with a given precision compared to naive Monte Carlo. A sample of naively drawn simulations can be found here: \url{https://youtu.be/6nuiLkecik8}.

\section{Case study}\label{sec:experiments}

We consider a highway scenario consisting of six agents, five of which are part of the environment. All vehicles are constrained to start within a set of possible initial configurations consisting of pose and velocity, and each vehicle has a goal of reaching the end of an approximately 1 kilometer stretch of road within CARLA's built-in ``Town04'' environment~\citep{dosovitskiy2017carla}. We also modify continuous weather parameters as well as the behaviors of the environment agents. See Appendix \ref{sec:appendix_p0} for a full description of the distribution $P_0$.  We cap simulations at 10 seconds in length (and they may be shorter if there is a collision). Throughout these experiments, we employ minimum time-to-collision (TTC) over a simulation (see Appendix \ref{sec:appendix_ttc} for a more detailed discussion) as our objective function $f(X)$. First, we compare the efficiency of our approach with naive Monte Carlo in generating statistically-confident estimates of rare-event probabilities. We then characterize the failure scenarios and their implications for \op's policy.

\subsection{Comparison with naive Monte Carlo}
\begin{figure}[t]
	\begin{minipage}{0.495\columnwidth}
		\centering
		\subfigure[95\% confidence intervals]{\includegraphics[width=0.99\textwidth]{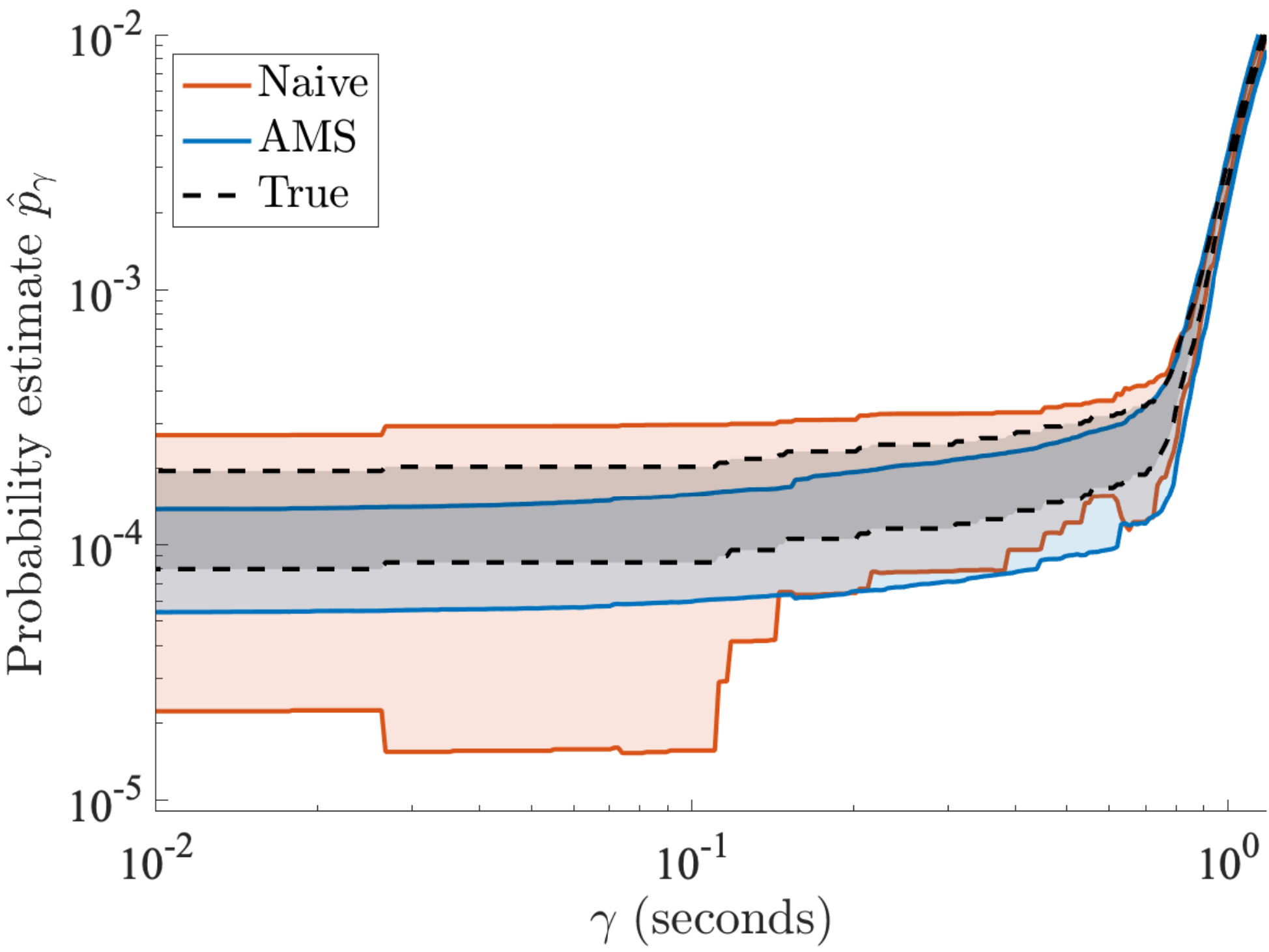}}%
	\end{minipage}
	\centering
	\begin{minipage}{0.495\columnwidth}%
		\centering
		\subfigure[Variance ratio]{\includegraphics[width=0.99\textwidth]{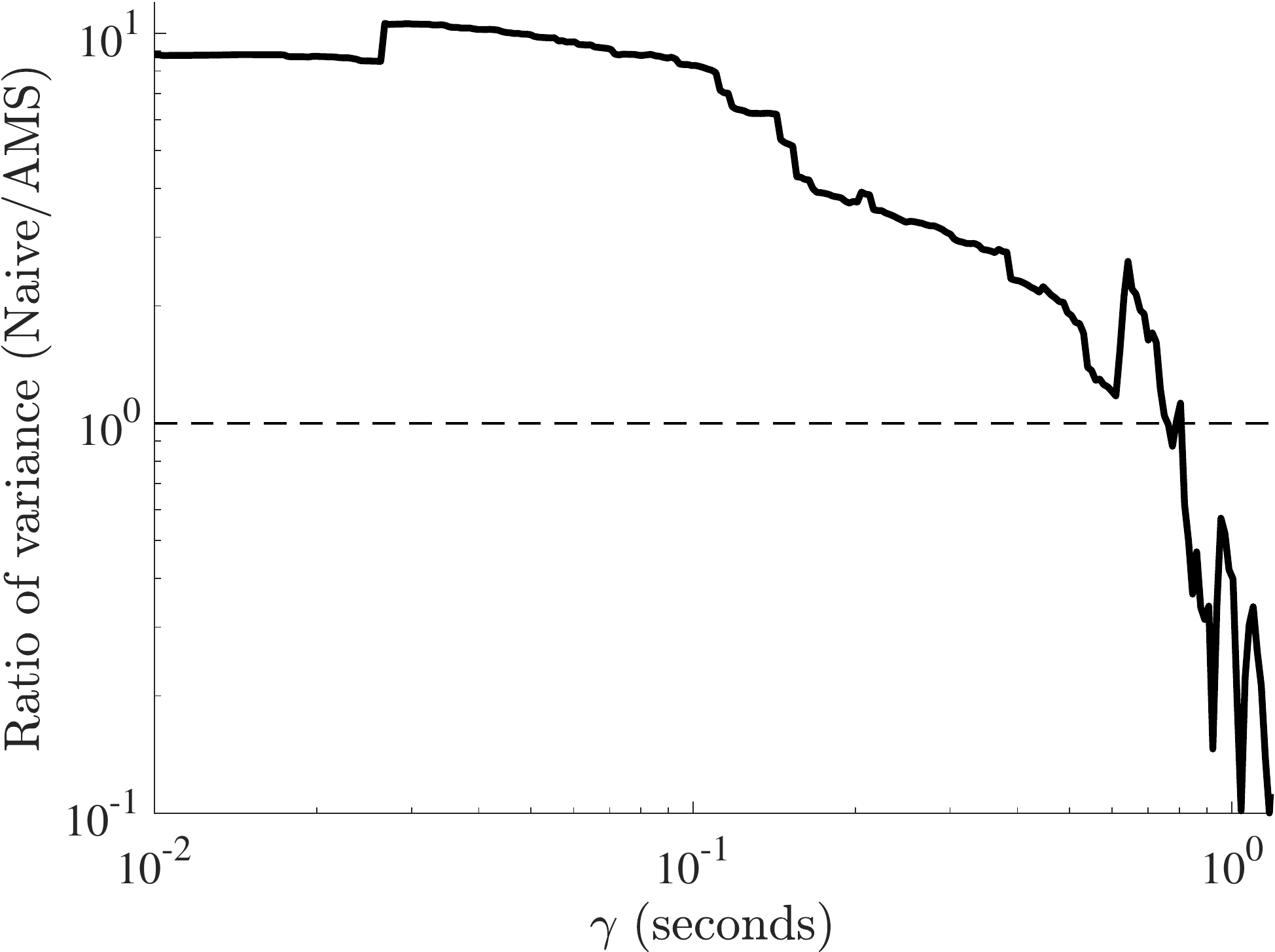}}%
	\end{minipage}
	\centering
	\caption[]{\label{fig:stuff}(a) 95\% confidence intervals for $\hat p_{\thresh}$ for 7 runs of AMS with 450 initial samples and naive Monte Carlo with the same number of simulation evaluations. `True' corresponds to naive Monte Carlo with 160,000 samples. (b) Ratio of the variance for naive Monte Carlo and AMS. As expected, boosts in performance by AMS increase with rarity ($\thresh$ smaller).}
\end{figure}

We begin by comparing the output of AMS (without the flow-based importance sampler) to naive Monte Carlo. Namely, we run AMS with a given threshold TTC value $\gamma$ and allow naive Monte Carlo to use same number of resulting samples (\eg~evaluations of the simulator). Figure~\ref{fig:stuff} compares performance of the two algorithms. In Figure~\ref{fig:stuff}(a), we show confidence intervals for the estimated probabilities along with a ground-truth value estimated using naive Monte Carlo with roughly 160,000 samples. In Figure~\ref{fig:stuff}(b), we plot the ratio of the variance of the two estimators. For events with probability lower than $10^{-3}$ ($\gamma \le 0.8$ seconds), AMS outperforms naive Monte Carlo, and the variance of the failure probability is reduced by up to $10\times$ (when $\gamma \le 0.1$ seconds). On the other hand, naive Monte Carlo is more efficient than AMS at estimating non-rare events ($\gamma > 0.8$ seconds). Interestingly, for this scenario and metric $f$, there is a steep exponential drop in likelihood for events near $\gamma \approx 1$ second, which happens to correspond with the regime below which our adaptive methods outperform naive Monte Carlo. 

\begin{figure}[h]
	\begin{minipage}{0.495\columnwidth}
		\centering
		\subfigure[Ratio of event frequency]{\includegraphics[width=0.99\textwidth]{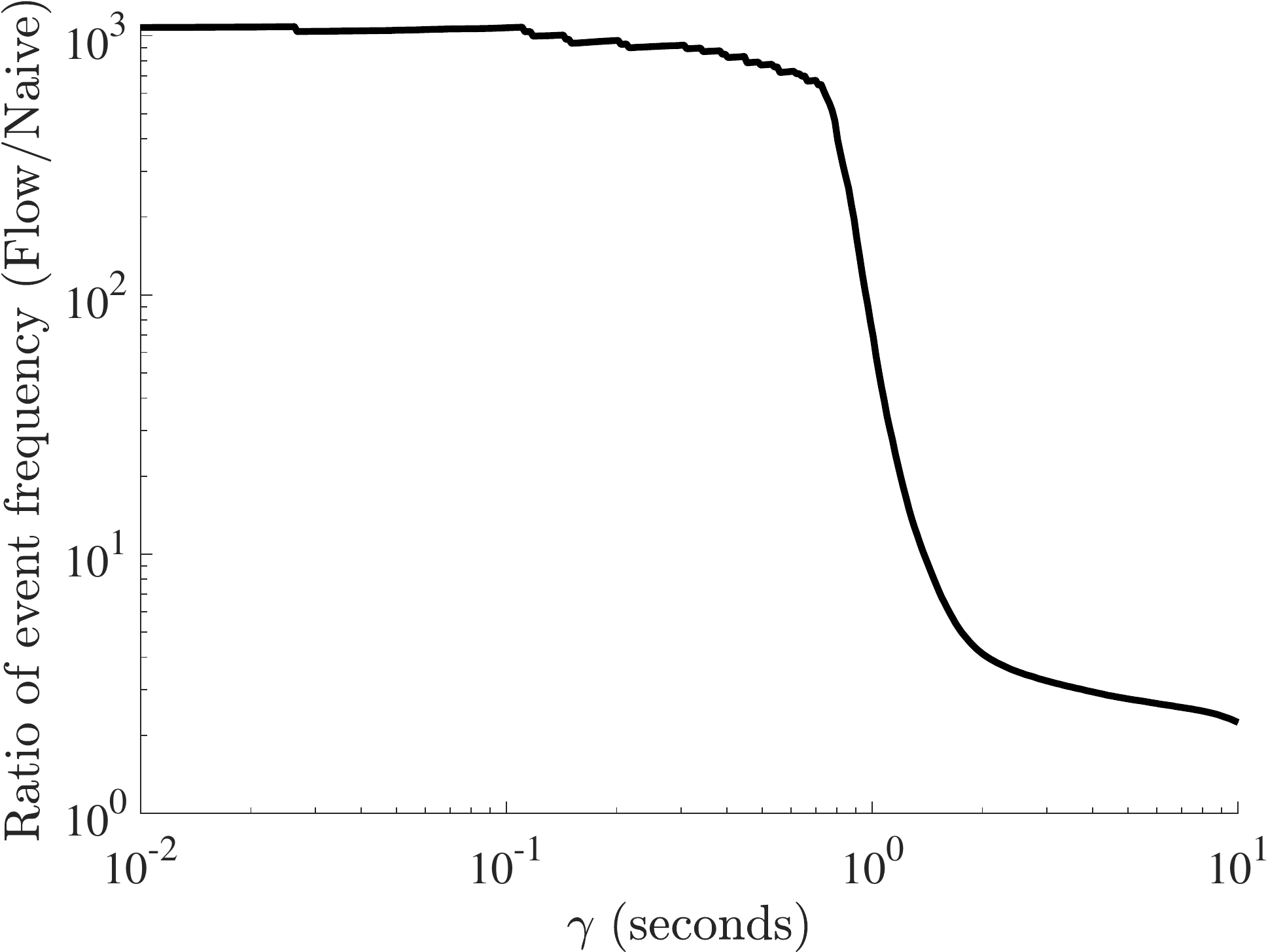}}%
	\end{minipage}
	\centering
	\begin{minipage}{0.495\columnwidth}%
		\centering
		\subfigure[Variance ratio]{\includegraphics[width=0.99\textwidth]{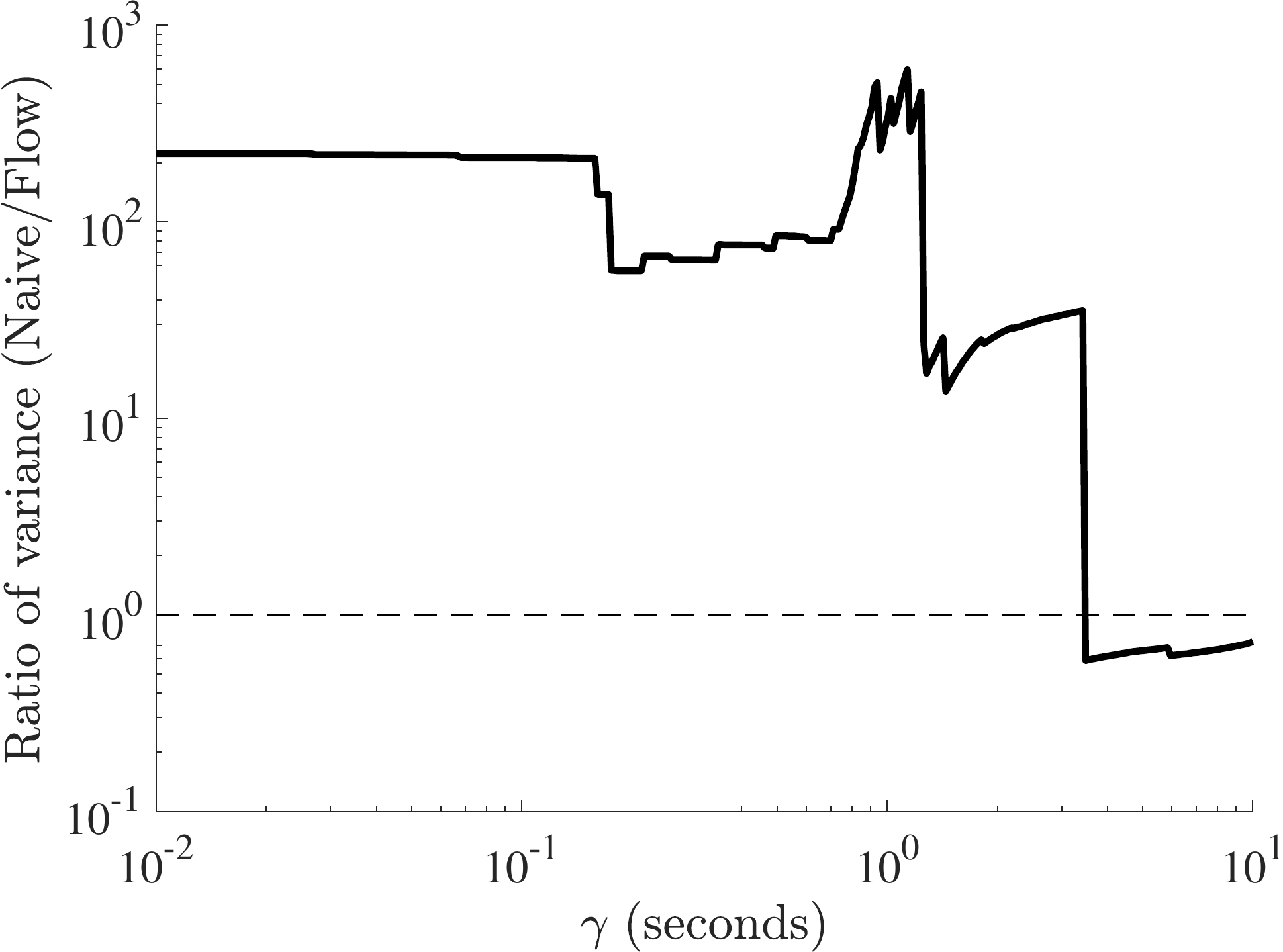}}%
	\end{minipage}
	\centering
	\caption[]{\label{fig:stuff2} Ratios for event frequency and variance of $\hat p_\gamma$ for flow-based importance sampler built upon AMS output vs. naive Monte Carlo sampling. The naive sampling budget is the number of samples drawn by the flow-based importance sampler \emph{plus} the total number of samples required by AMS.} %
\end{figure}

Figure~\ref{fig:stuff2} shows results from the flow-based importance sampler built upon the output of AMS. Importantly, to compare the performance of this sampler properly with that of naive Monte Carlo, we allow naive Monte Carlo to have a number of samples equal to the amount sampled by the flow \emph{in addition} to the amount required to run AMS. In Figure~\ref{fig:stuff2}(a), we illustrate the fact that the flow-based importance sampler places high likelihood in failure regions, sampling three orders of magnitude more failures than naive Monte Carlo for very dangerous events ($\gamma \le 0.1$ seconds). In Figure~\ref{fig:stuff2}(b) we see that the importance-sampling estimate has variance two orders of magnitude smaller than that of naive Monte Carlo for events with $\gamma \le 1$ second. In other words, naive Monte Carlo requires two orders of magnitude more simulations than our approach to generate estimates of failure probabilities with the same level of statistical confidence. 

\subsection{Analysis of failure probability and failure modes}
\paragraph*{Failure probability}
We choose $\gamma=0.1$ seconds as our threshold for failure, so that we are interested in the probability of events with TTC $< 0.1$ seconds. This threshold encompasses both situations where there is no actual crash but the event is still extremely dangerous as well as situations that result in actual impact. As illustrated in Figure \ref{fig:stuff}(a), the failure rate is roughly 1 in $10^4$ simulations, which, given the average speeds of the vehicles (roughly 20 meters per second) and the simulation length (10 seconds), corresponds to 1 in 1250 miles.

The only reference values with which to compare this estimated failure rate are the reported disengagement rates by AV manufacturers, but we do so with strong caveats. Disengagements are defined as situations where a human safety driver takes control of a vehicle that was operating autonomously. Importantly, there is no accepted standard for when a disengagement should occur nor do AV manufacturers publish their internal definitions. In 2018, Comma AI claimed a rate of 1 disengagement per hour of driving~\citep{bigelow18after}. Other manufacturers reported a slightly more transparent metric: Waymo claimed a disengagement rate of 1 per 11000 miles, Cruise claimed 1 per 5200 miles, Zoox claimed 1 per 1900 miles, and Nuro claimed 1 per 1000 miles~\citep{crowe18waymo}.

\begin{figure}[!!t]
\centering
	\begin{minipage}{0.45\columnwidth}
		\centering
		\subfigure[Failure clusters]{\includegraphics[width=0.9\textwidth]{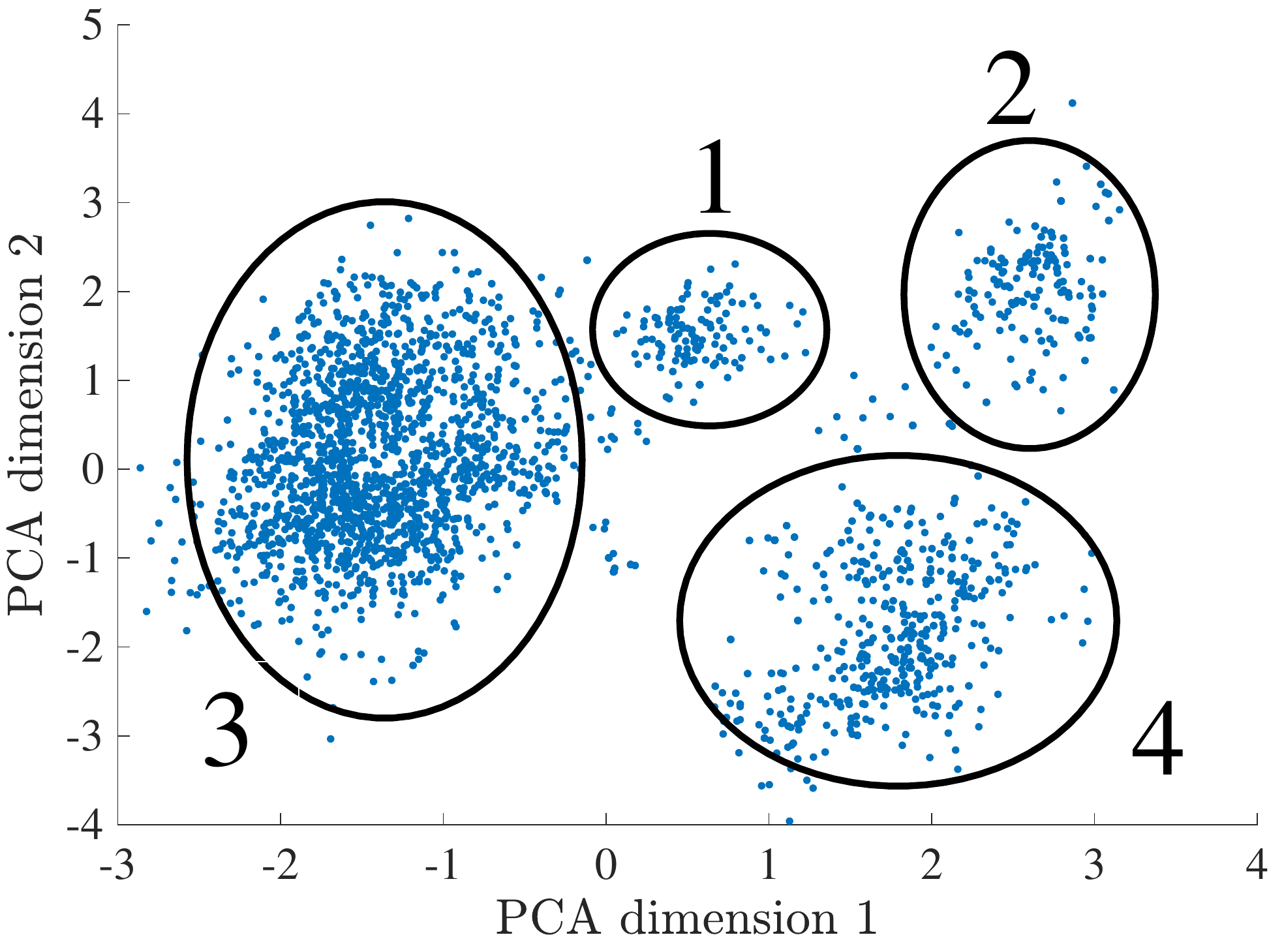}}%
	\end{minipage}
	
	\centering
	\begin{minipage}{0.245\columnwidth}%
		\centering
		\subfigure[Sun - Dashcam]{\includegraphics[width=1.0\textwidth]{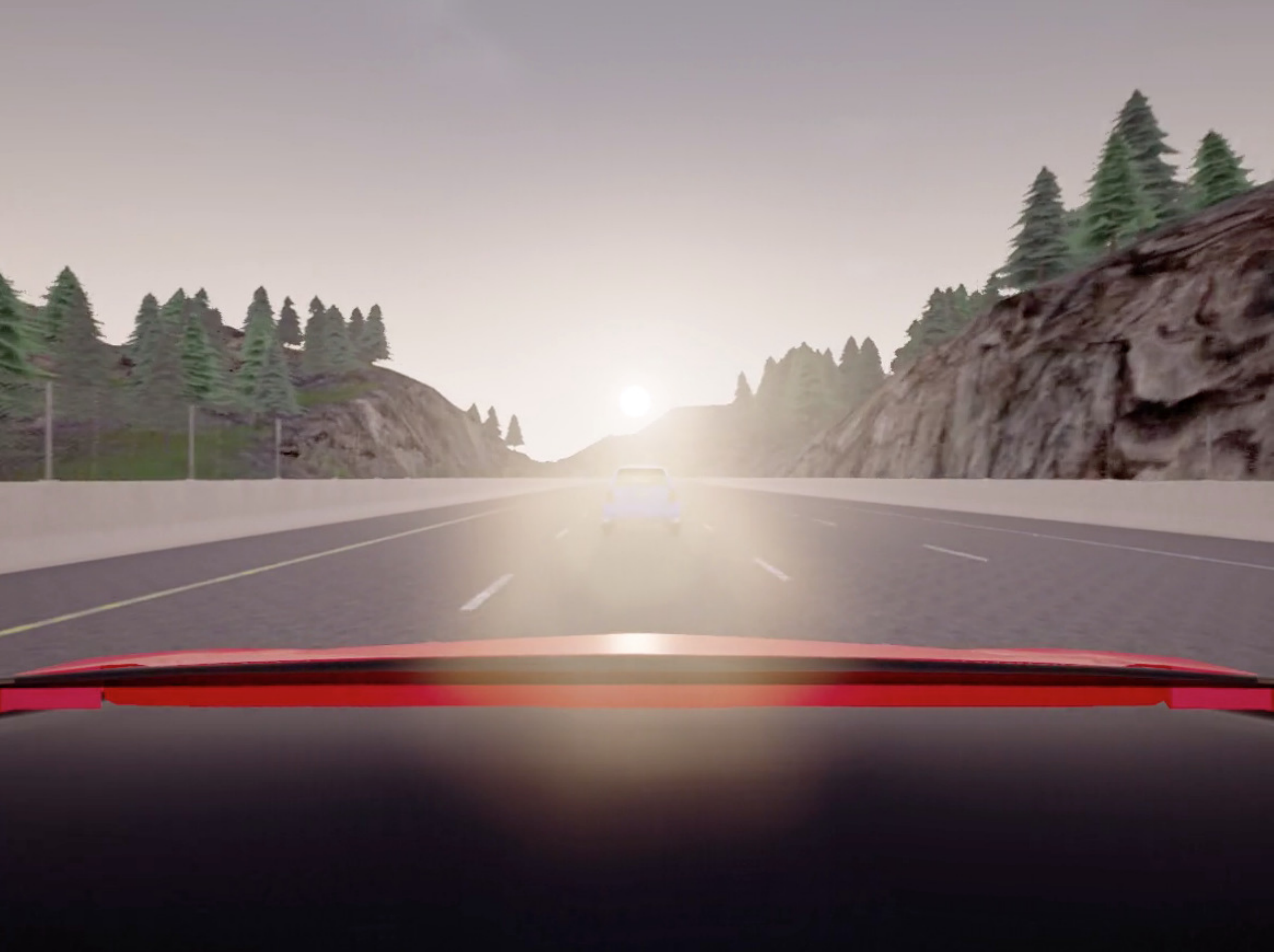}}%
	\end{minipage}
	\centering
		\begin{minipage}{0.245\columnwidth}%
		\centering
		\subfigure[Sun - Overhead]{\includegraphics[width=1.0\textwidth,height=3.05cm]{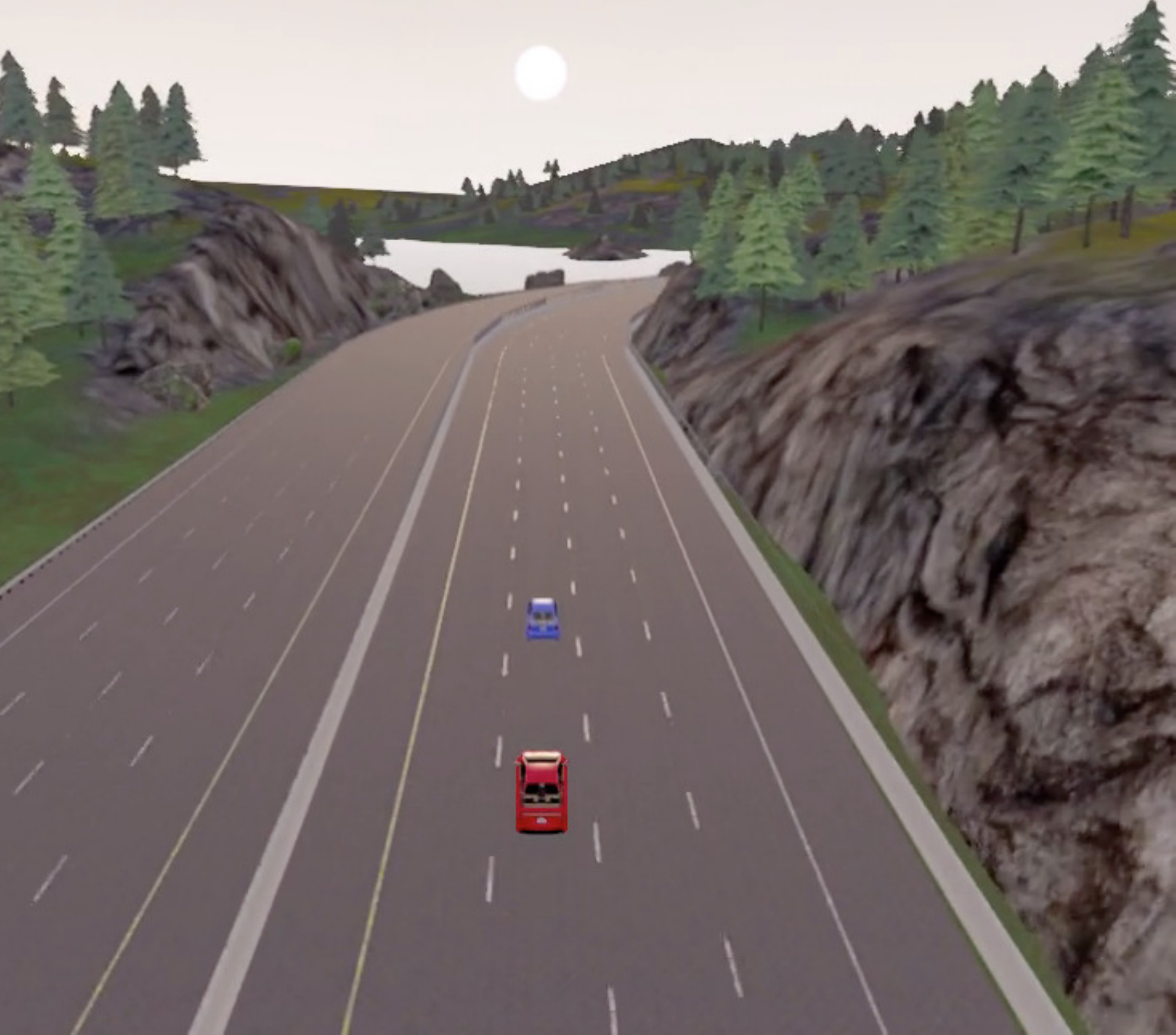}}%
	\end{minipage}
		\centering
	\begin{minipage}{0.245\columnwidth}%
		\centering
		\subfigure[Rain - Dashcam]{\includegraphics[width=1.0\textwidth,height=3.05cm]{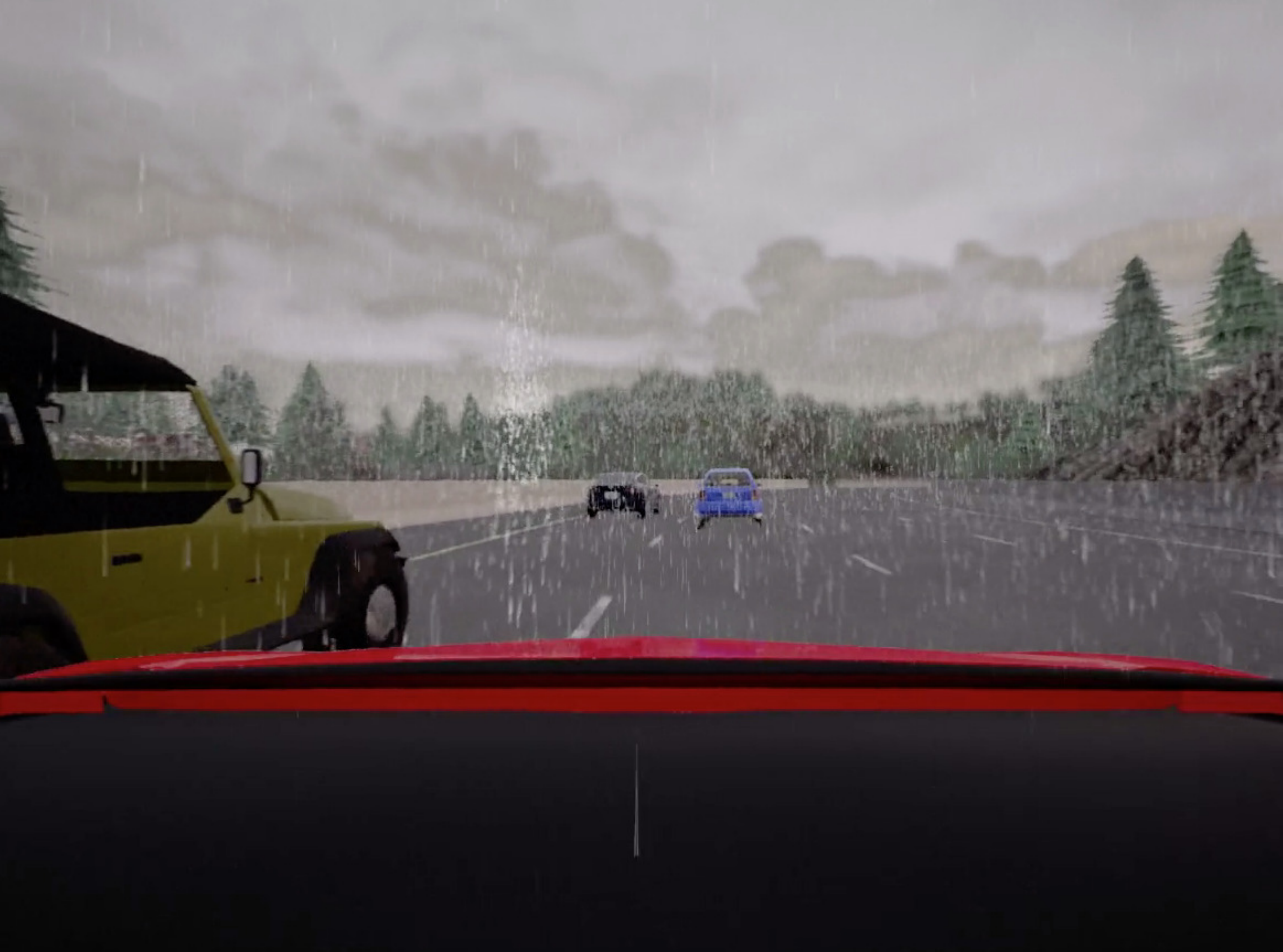}}%
	\end{minipage}
	\centering
		\begin{minipage}{0.245\columnwidth}%
		\centering
		\subfigure[Rain - Overhead]{\includegraphics[width=1.0\textwidth,height=3cm]{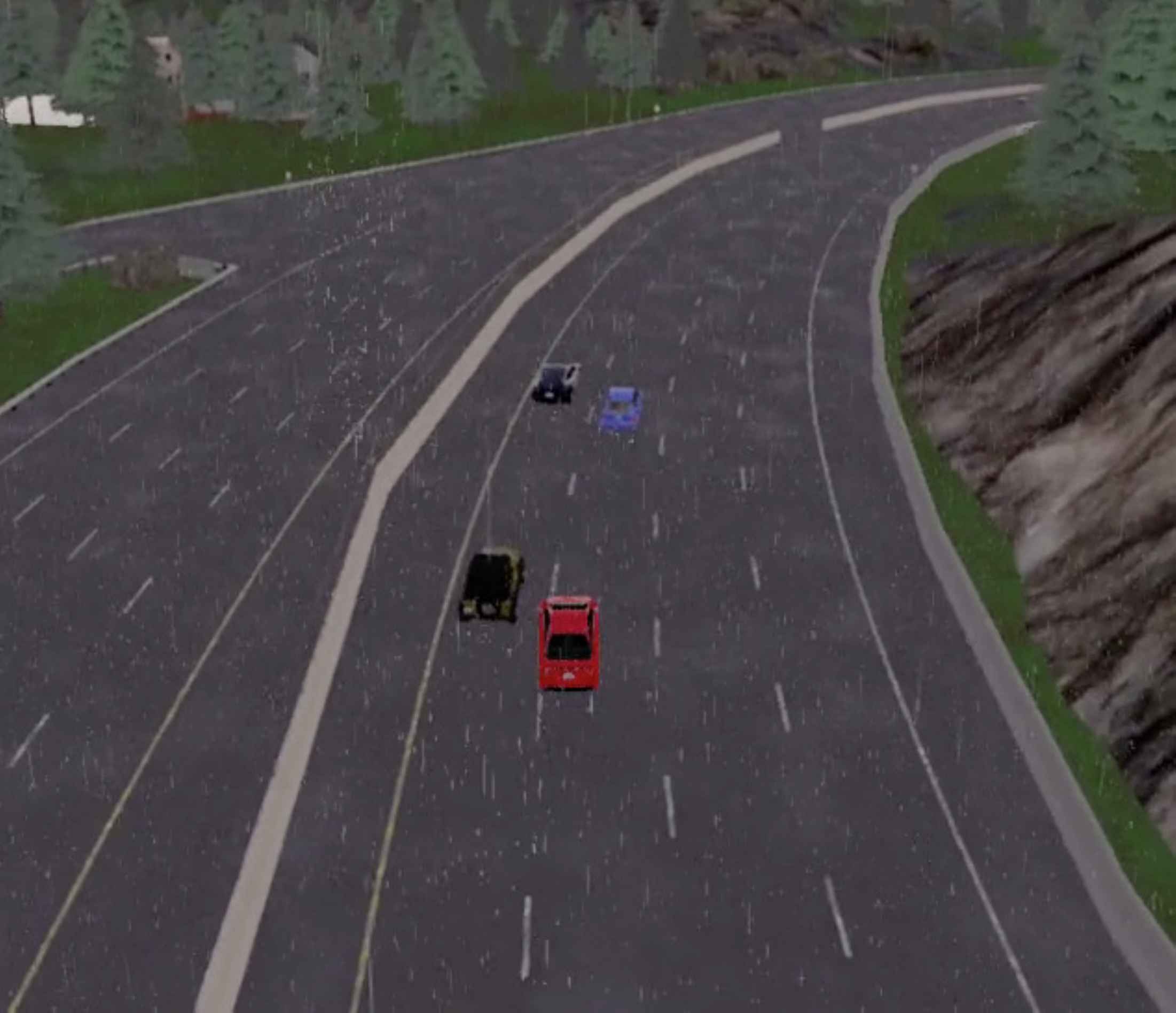}}%
	\end{minipage}
	\caption[]{\label{fig:PCA} Analysis of failure modes. (a) Graphical depiction of failure modes in 2 dimensions via PCA. (b)-(c) Views from a crash simulation where direct sunlight caused errors in the deep-learning-based vision system, and {\op} failed to see the blue lead car and rear-ended it. (d)-(e) Views from a crash simulation where rain caused errors in the vision system and {\op} misjudged lane boundaries. A playlist of failure mode examples can be viewed at \url{https://www.youtube.com/playlist?list=PL4lTzImUFh3xhx7kvjrIh6EZudg2n5R3C}.}
\end{figure}

\paragraph*{Failure modes}
Since we have access to the {\op} codebase, we can investigate the failure modes. An elementary linear dimensionality reduction of the failure conditions $X$ into 2 dimensions (via PCA) yields 4 visible clusters, as depicted in Figure~\ref{fig:PCA}(a). Whereas the first PCA mode largely characterizes changes in weather conditions  (\eg~sunlight in clusters 1 and 3 vs. rain in clusters 2 and 4), the second PCA mode characterizes changes in relative velocities between vehicles (low relative velocity in clusters 1 and 2 vs. high relative velocity in clusters 3 and 4). At first glance, the predominant culprit of failures appears to be the perception system (but we investigate this claim further below); it has high uncertainty about vehicles ahead of it as well as lane boundaries when there is direct sunlight on the camera or heavy rain (Figures~\ref{fig:PCA}(b)-(c) and \ref{fig:PCA}(d)-(e)). The crashes are either glancing collisions when relative velocities are low or high-impulse collisions when relative velocities are high. Furthermore, the median likelihood of failure due to precipitation is two times higher than that due to direct sunlight using our generative model. The implications of these failures for potential improvements in \op~are clear: (a) the vision system can be improved to better handle these adverse weather conditions, and (b) the feedback loop between the vision and the control stacks can be made more conservative in the presence of high uncertainty. 

\begin{figure}[!t]
	\begin{minipage}{0.49\textwidth}
		\centering
		\subfigure[Safe behavior]{\includegraphics[width=1.0\textwidth]{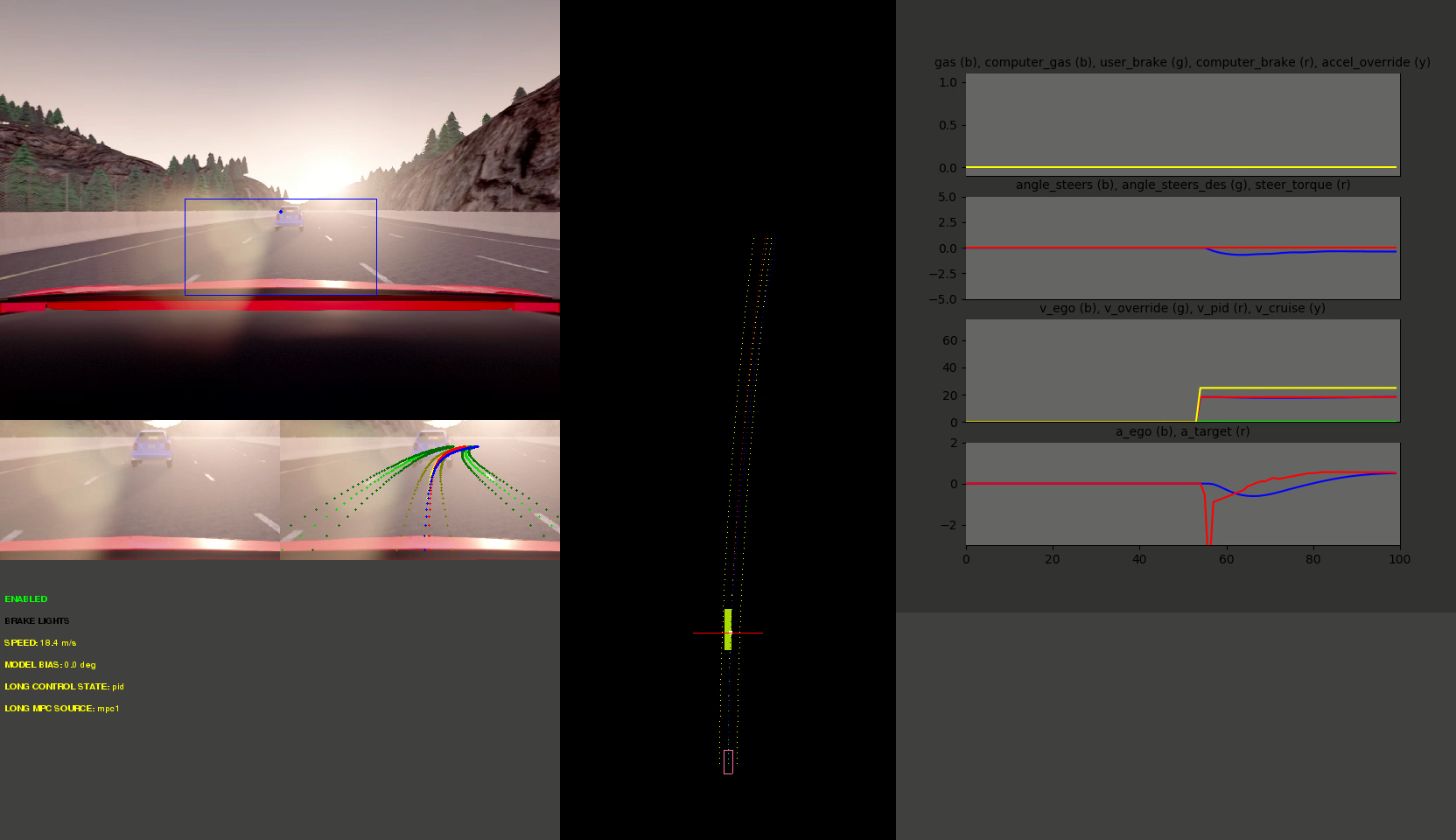}}%
	\end{minipage}
	\centering
	\begin{minipage}{0.49\textwidth}%
		\centering
		\subfigure[Unsafe behavior]{\includegraphics[width=1.0\textwidth]{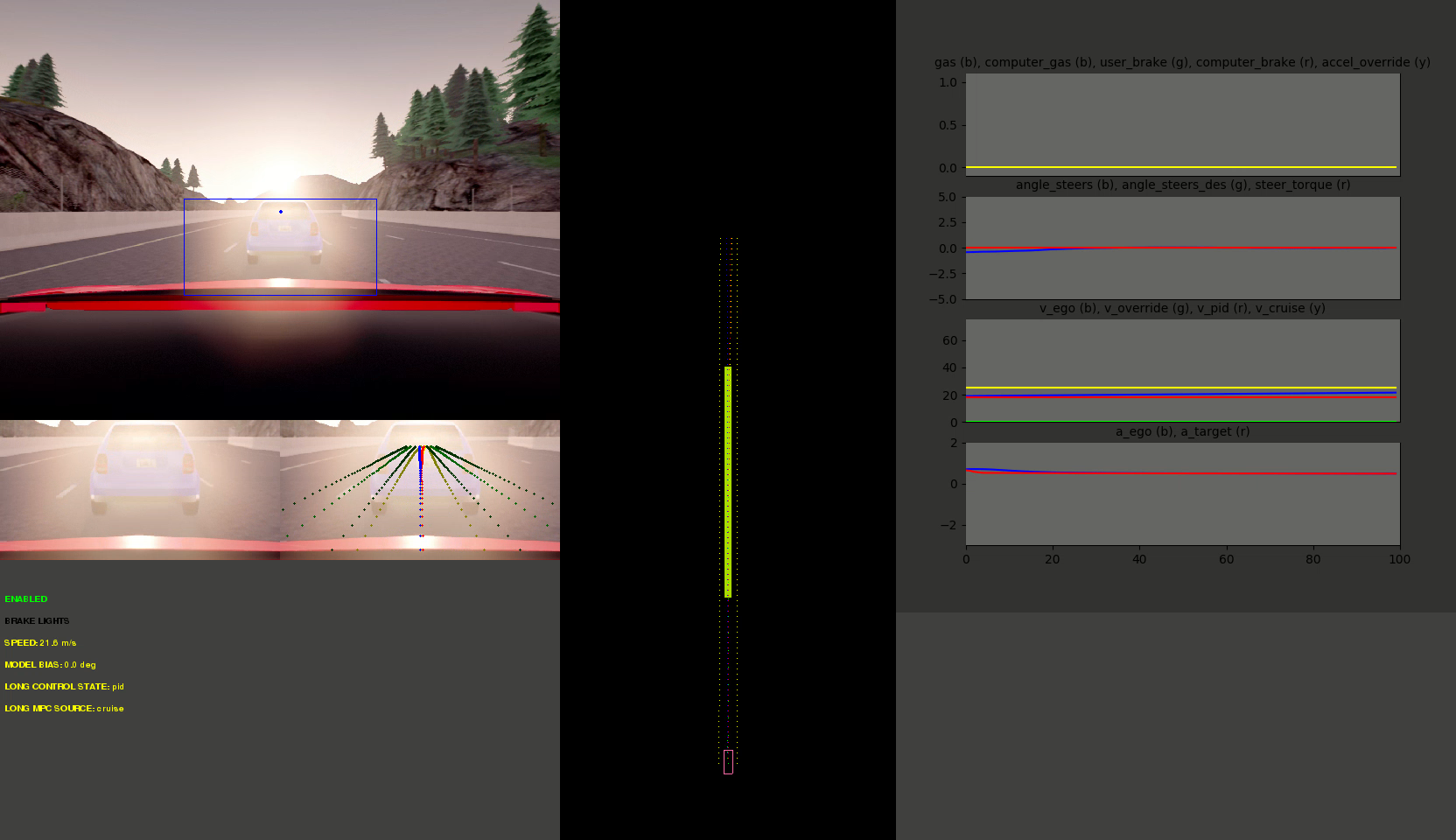}}%
	\end{minipage}
	\caption{Failure at the interface between perception and planning. (a) The perception system has low uncertainty of the lead car's location. (b) The blinded perception system places large uncertainty on the lead car's location. This uncertainty is not used well by the planning module and the ego-vehicle accelerates into the lead vehicle.}
	\label{fig:crash}
	\vspace{-5pt}
\end{figure}

We highlight the importance of the latter potential improvement in the feedback between perception and planning. Namely, although the vision system is clearly a cause of these failures, it is not the \emph{sole} cause. Rather, these failure modes (particularly the modes that involve direct sunlight) lie at the interface between the perception and planning systems. Figure \ref{fig:crash} substantiates this claim by viewing the {\op} debug output during one of the crash scenarios.  In Figure \ref{fig:crash}(a), the ego-vehicle, not yet blinded by the sun, notices the lead car as well as the lane boundaries (left side of image). The location of the lead vehicle has low uncertainty (middle of image). In Figure \ref{fig:crash}(b) the ego-vehicle becomes blinded by the sun, and the perception system correctly interprets that it is very uncertain about the position of the lead vehicle. The MPC is given the estimated pose of the lead vehicle and plans a safe trajectory relative to its knowledge of the world. Unfortunately, the constraints in the MPC don't treat the uncertainty of the estimated pose in a conservative manner (namely the mean of the estimated pose gets shifted forward substantially), allowing the ego-vehicle to accelerate into the lead vehicle.  Importantly, neither the perception nor planning system fails catastrophically in this example, and this type of integrated system failure cannot be found by investigating either system in isolation.

\section{Conclusion}
Our risk-based framework, combined with efficient adaptive search in black-box simulation, is an approach to efficiently quantify risk and achieve scalable CI for AVs.
Real-world testing is still the (slow, expensive, and dangerous) gold standard for asserting a given level of safety for an AV policy, and it is best applied to confirm rather than replace evidence from simulations. Verified subsystems also provide evidence that components meet their design specifications. However, verification is limited by computational intractability, the quality of the specification, and access to white-box models.
Rule-based systems for evaluating failures do not yield tangible or actionable safety results, whereas risk-based models for evaluating AV policies are both more objective and actionable. Additionally, while obvious adversarial ``corner cases'' are sometimes evocative, the failures that need prioritization are those in the seemingly benign long-tail of less-exotic but more-frequent failure situations.
Automating the discovery of these less-exotic yet more-frequent failures is  paramount to understanding, deploying, and improving AV systems beyond human-level performance.

The risk-based framework is most useful when the base distribution $P_0$ is accurate, and otherwise it can suffer from systemic biases. Namely, the unbiasedness of the search algorithm is conditioned on the search space---no search algorithm can inherently correct for a poor simulation environment that has an unrealistic rendering engine or unrealistic models of human behavior. Nevertheless, although an estimate of $p_\gamma$ suffers from bias when $P_0$ is misspecified, our adaptive sampling techniques still efficiently identify dangerous scenarios which are independent of potentially subjective assignments of blame. Principled techniques for building and validating the model of the environment $P_0$ represent open research questions.

Importantly, the risk-based framework can easily be combined with existing approaches to testing in simulation. The family of nonparametric importance-sampling algorithms outlined in Section \ref{sec:AIS} is a drag-and-drop replacement for existing methods that perform grid search, fuzzing, or naive Monte Carlo search over a parameter space. Furthermore, metrics that assign blame, such as those outlined by~\citet{shalev2017formal}, can be incorporated into the objective function $f$ and used to guide the search for failures. Finally, even though it is designed to work with the whole black-box system, our risk-based framework can be applied to search for failures in subsystems (\eg~the post-perception planning module) as well.

Overall, our framework offers significant speedups over real-world testing, allows efficient, automated, and unbiased analysis of AV behavior, and ultimately provides a powerful tool to make safety analysis a tractable component in AV design. We believe that rigorous safety evaluation of AVs necessitates adaptive benchmarks that maintain system integrity and prioritize system failures. The framework presented in this paper accomplishes these goals by utilizing adaptive importance-sampling methods which require only black-box access to the policy and testing environment. As demonstrated on Comma AI's {\op}, our approach provides a principled, statistically-grounded method to continuously improve AV software. More broadly, we believe this methodology can enable the feedback loop between AV manufacturers, regulators, and insurers that is required to make AVs a reality.

\setlength{\bibsep}{0.5pt}
\bibliographystyle{abbrvnat}

\newpage

\appendix

\section{Simulation parameters}\label{sec:appendix_p0}
The 1 kilometer stretch of road that we consider consists of a straight section followed by a right-turning bend and a final straight section. Our search space $P_0$ consists of weather parameters, initial positions and velocities of the vehicles, and the behaviors of the environmental agents.

4 parameters of the weather are varied: precipitation on the ground ($P_g$), the altitude angle of the sun ($A$), cloudiness ($C$), and precipitation in the air ($P_a$). Precipitation on the ground follows the distribution $P_g\sim 50\text{Uniform}(0,1)$, where the units are proprietary for the CARLA simulator. Sun altitude follows the distribution $A\sim 90\text{Uniform}(0,1)$, whose units are in degrees. Letting  $C_b\sim\text{Beta}(2,2)$ and $C_u\sim\text{Uniform}(0,1)$, cloudiness has a distribution given by the following mixture:
\begin{equation*}
C \sim 30C_b \mathbf{1}\{C_u < 0.5\} + (40C_b+60)\mathbf{1}\{C_u \ge 0.5\},
\end{equation*}
where $\mathbf{1}\{\cdot\}$ is the indicator function and the units of $C$ are in proprietary units for CARLA. Precipitation in the air is a deterministic function of cloudiness, $P_a = C\mathbf{1}\{C \ge 70\},$ where the units are again proprietary CARLA units.

Initial positions for each vehicle are independently given by the pair $(S,T)$, where $S\sim 500 \text{Beta}(2,2) + 200$ denotes the longitudinal position along the stretch of road with respect to the start of the road, and $T\sim0.5 \text{Beta}(2,2)-0.25$ denotes the lateral distance from the lane center with left being positive. Both have units in meters. Initial velocities are independently given by $V \sim 10 \text{Beta}(2,2)+15$ with units in meters/second.

For the behaviors of the environment vehicles we follow the parametrization of \citet{okelly2018scalable}. Namely, we search over the last layer of a neural network GAIL generator model. We use $\xi \in \R^{404}$ to denote the weights of the
last layer of the neural network trained to imitate human-like driving
behavior with $\xi \sim \mathcal{N}(\mu_0, \Sigma_0)$, with the mean and covariance matrices learned via empirical bootstrap. The input for this GAIL generator is described in \citet{okelly2018scalable}. The single network defined by $\xi$ drives the behavior for all environment agents in a particular simulation rollout.

Together, the weather, position, velocity, and behavior variables form samples $X\in \R^{426}$.

\section{TTC calculation}\label{sec:appendix_ttc}
Let $T_i(t)$ be the instantaneous time-to-collision between the ego vehicle and the $i$-th environment vehicle at time step $t$. For a given simulation rollout $X$, the TTC metric is given by
\begin{equation*}
f(X):=\min_{t} \left ( \min_{i}T_i(t)  \right ).
\end{equation*}
The value $T_i(t)$ can be defined in multiple ways (see~\eg~\citet{sontges2018worst}). In this study, we define it as the amount of time that would elapse before the two vehicles' bounding boxes intersect assuming that they travel at constant fixed velocities from the snapshot at time $t$.

\end{document}